# Credit Card Fraud Detection Using RoFormer Model With Relative Distance Rotating Encoding

Kevin Reyes
*Flow Payment Gateway*
Santiago, Chile.
kreyes@flow.cl

Vasco Cortez
*Flow Payment Gateway*
Santiago, Chile
vcortez@flow.cl

*Abstract*— Fraud detection is one of the most important challenges that financial systems must address. Detecting fraudulent transactions is critical for payment gateway companies like Flow Payment, which process millions of transactions monthly and require robust security measures to mitigate financial risks. Increasing transaction authorization rates while reducing fraud is essential for providing a good user experience and building a sustainable business. For this reason, discovering novel and improved methods to detect fraud requires continuous research and investment for any company that wants to succeed in this industry. In this work, we introduced a novel method for detecting transactional fraud by incorporating the Relative Distance Rotating Encoding (ReDRE) in the RoFormer model. The incorporation of angle rotation using ReDRE enhances the characterization of time series data within a Transformer, leading to improved fraud detection by better capturing temporal dependencies and event relationships.

## I. INTRODUCTION

Fraud detection and prevention has been widely studied, especially since the 2000s, when large companies like eBay and Amazon experienced rapid growth due to increased internet penetration. In recent years, this field has become indispensable, driven by the significant expansion of e-commerce, which now accounts for approximately 20% of total retail sales worldwide [1]. The growing complexity and scale of online transactions, coupled with the sophistication of fraudulent schemes, have underscored the need for advanced fraud detection mechanisms. This evolution has led to the development of innovative models and algorithms to protect consumers and businesses from financial losses and maintain trust in digital commerce ecosystems [2-4].

While fraud detection and prevention may seem synonymous, they refer to distinct concepts. Fraud prevention involves identifying and stopping fraudulent activity before it happens, whereas fraud detection refers to identifying fraud only after it has occurred [5]. Usually, the model should be tested and proven in fraud detection using historical data or public datasets, to later be incorporated into fraud prevention systems that allow transactions to be declined in real or near-real-time prevention system. The Fraud detection is one of the most important challenges that financial systems must address. Detecting fraudulent transactions is critical for payment gateway companies, which process millions of transactions monthly and require robust security measures to mitigate financial risks.

Fraud systems detect fraudulent or abnormal patterns from previous observations. They analyze different transaction information, such as IP locations, sign-up times, purchase items, user IDs, device IDs, emails, websites, among others. This process is conceptually similar to cybersecurity attack detection, where security systems monitor network traffic for signs of intrusion. In both cases, the challenge lies in distinguishing between benign and malicious activity while minimizing false positives. Transactional fraud detection, like cyber threat detection, must evolve continuously to counter increasingly sophisticated attack methods.

There are different public databases used to test fraud detection models. One of the most popular ones, and the one we used for this work, is the Fraud Dataset Benchmark (FDB), which is a compilation of publicly available datasets relevant to fraud detection [6]. The dataset source used in this work is *IEEE-CIS Fraud Detection*, also called "ieeecis". Prepared by IEEE Computational Intelligence Society, this card-non-present transaction fraud dataset was launched during IEEE-CIS Fraud Detection Kaggle competition and was provided by Vesta Corporation [7].

The RoFormer (Rotary Transformer) model [8], developed in 2023, introduces a novel approach to handling positional information in sequences through Rotary Position Embeddings (RoPE) in the Transformer model. This method builds upon the widely adopted Transformer model [9], which has become a cornerstone in Artificial Intelligence tools for tasks such as natural language processing, computer vision, and time-series analysis. In this work, we introduced a variation in the RoFormer model, Relative Distance Rotating Encoding (ReDRE), which differs from RoFormer in how it characterizes the distance between events. In ReDRE, the distance is relative and can vary depending on the temporality of the events, whereas in the RoFormer, the distance between events is equispaced.

The RoFormer and its variant, ReDRE, may enhance the efficiency and scalability of the Transformer model, making it particularly effective for long-sequence tasks and improving performance across a variety of AI applications. The introduction of a geometric property into the embedding vectors allows the Transformer to better capture sequential and temporal relationships within events. Relative positional dependencies are critical, such as detecting fraudulent transactions over extended periods.

By its nature, transactional data is a temporal representation of a series of activities performed by a payer or cardholder over a specific period. While the Transformer model relies on positional encodings to represent order, the inherent temporal characteristics of the transactions are not directly captured in this framework. This limitation can make it challenging for standard Transformer architectures to fully leverage the irregularly spaced sequential nature with non-uniform spacing of transactional data.

This paper presents a novel mechanism for detecting transactional fraud by incorporating ReDRE into the RoFormer model, enhancing its ability to capture dynamic temporal dependencies. First, we provide a detailed explanation of both the RoFormer and RoFormer ReDRE models, highlighting their key differences and advantages. Then, we evaluate their performance alongside the classic Transformer model using the IEEE-CIS Fraud Detection dataset, analyzing their effectiveness by calculating the Area Under the Receiver Operating Characteristic (AUC-ROC) curve. The AUC-ROC curve allows us to assess how well the model detects fraudulent transactions.

## II. METHOD

We shall consider a token as a unique credit card transaction, where the sequence $\{w_i\}_{i=1}^{N}$ corresponds to N transactions carried out during the period. The corresponding feature embedding is denoted as $\mathbb{E}_N = \{x_i\}_{i=1}^{N}$ where $x_i \in \mathbb{R}^d$ is the d-dimensional feature embedding of token $w_i$.

The self-attention mechanism first incorporates positional information into the embeddings vector, enriching them with context about their order within the sequence. These enriched embeddings are then transformed into query ($q$), key ($k$), and value ($v$) representations.

$$q_m = f_q(x_m, m),$$
$$k_n = f_k(x_n, n),$$
$$v_n = f_v(x_n, n)$$

where $q_m$, $k_n$ and $v_n$ incorporate the $n^{th}$ and $m^{th}$ positions through $f_q$, $f_k$ and $f_v$, respectively.

The attention mechanism calculates the relationships between tokens by computing the dot product between $q_m$ and $k_n$, weighting the value vectors $v_n$ based on the resulting attention scores.

### A. RoFormer

The RoFormer incorporates the Rotary Position Embedding (RoPE) in the Transformer. It is an approach to incorporate positional information directly into the projections of queries $q_m$ and keys $k_n$. This method uses a rotational representation in a complex space to preserve the relative properties of positions.

For a position matrix encoded via RoPE, a rotation is applied to the $q_m$ and $k_n$ vectors. The $v_n$ vector does not explicitly incorporate positional information, as this only affects the comparison between $q_m$ and $k_n$ during the attention calculation.

It can be represented the queries and keys vectors with the rotary position as follows:

$$q'_m(p) = R(p) \cdot q_m,$$
$$k'_n(p) = R(p) \cdot k_n$$

Where $q'_m$ and $k'_n$ vectors are the position-encoded queries and keys vectors, and $R(p)$ is the rotary matrix that encodes positional information $p$.

### B. Relative Distance Rotating Encoding

The Relative Distance Rotating Encoding (ReDRE) introduces a rotation angle based on an explicit distance between two events in the series which in this case corresponds to the temporal distance between the transaction at the position $k$ and the transaction under evaluation.

$$\theta_k = d \cdot \omega_k$$

Where $d$ is the relative distance between the event of interest and the current event.

### C. General Form of Rotary Matrix

For a given embedding vector $q \in \mathbb{R}^d$, where $d$ is the dimensionality of the embedding, the rotation matrix $R(P)$ is a block-diagonal matrix that rotates each pair of dimensions independently based on the positional index $p$.

$$R(p) = \begin{bmatrix} \cos(\theta_1) & -\sin(\theta_1) & 0 & 0 & \cdots & 0 & 0 \\ \sin(\theta_1) & \cos(\theta_1) & 0 & 0 & \cdots & 0 & 0 \\ 0 & 0 & \cos(\theta_2) & -\sin(\theta_2) & \cdots & 0 & 0 \\ 0 & 0 & \sin(\theta_2) & \cos(\theta_2) & \cdots & 0 & 0 \\ \vdots & \vdots & \vdots & \vdots & \ddots & \vdots & \vdots \\ 0 & 0 & 0 & 0 & \cdots & \cos(\theta_{d/2}) & -\sin(\theta_{d/2}) \\ 0 & 0 & 0 & 0 & \cdots & \sin(\theta_{d/2}) & \cos(\theta_{d/2}) \end{bmatrix}$$

Each pair of dimensions $(i, i+1)$ is transformed using a $2x2$ submatrix:

$$R(\theta_k) = \begin{bmatrix} \cos(\theta_k) & -\sin(\theta_k) \\ \sin(\theta_k) & \cos(\theta_k) \end{bmatrix}$$

### D. General Application to a Vector

If the embedding vector $q = [q_1, q_2, \ldots, q_d]$ is partitioned into pairs $(q_{2i}, q_{2i+1})$, the rotary transformation for each pair is:

$$\begin{bmatrix} q'_{2i} \\ q'_{2i+1} \end{bmatrix} = R(\theta_i) \cdot \begin{bmatrix} q_{2i} \\ q_{2i+1} \end{bmatrix}$$

Expanded, this becomes the general vectors:

$$q'_{2i} = \cos(\theta_i) \cdot q_{2i} - \sin(\theta_i) \cdot q_{2i+1},$$
$$q'_{2i+1} = \sin(\theta_i) \cdot q_{2i} + \cos(\theta_i) \cdot q_{2i+1}$$

In this work, we considered three models for our analysis. The first model used is the classic Transformer, which does not include a rotary matrix. The second model is the RoFormer, and the last one is the RoFormer with the ReDRE variation.

## III. EXPERIMENTAL EVALUATION

### A. Dataset

The dataset is composed of two major components: Transaction data and Identity data. The Transaction table encompasses both money transfers and goods/services transactions. The field TransactionDT represents a time delta in seconds from a reference datetime, spanning approximately six months. TransactionAMT indicates payment amounts in USD. ProductCD specifies the product or service type, whereas card1–card6 capture payment card attributes (e.g., card type, issuing bank, or country). The addr and dist fields store billing region/country and various distance metrics, while P_ and R_ refer to purchaser and recipient email domains. Only ~3.5% of the transactions in the dataset are labeled as fraudulent, which reflects the high-class imbalance commonly observed in real-world fraud detection scenarios.

Additional features offer contextual insights. C1–C14 are counting variables that record frequencies of addresses or devices used with a given payment card. D1–D15 represent time intervals between previous events, M1–M9 reflect matches across fields (e.g., name on card vs. billing name), and Vxxx are advanced features engineered by Vesta, often incorporating frequency or rank-based measures (e.g., the number of times a specific card–IP–email combination appears within a given time window). The Identity table covers network and device data, including IP addresses, ISP/proxy information, and digital fingerprints (e.g., browser user-agent and operating system).

### B. The experiment

A set of experiments was conducted to compare three different architectures for fraud detection: (1) a baseline traditional Transformer applied to time-series data, (2) a RoFormer variant that employs rotary position embeddings based on absolute positions, and (3) a modified RoFormer that incorporates a relative time-based rotation mechanism ReDRE. Each model was trained and evaluated under the same conditions, allowing for a direct assessment of their performance on the dataset.

All experiments used the same training and validation splits described in the dataset section. Class imbalance was addressed by weighting the positive class in the binary cross-entropy loss function, using the ratio of non-fraud to fraud samples. This ensured that the minority class had appropriate influence during training despite its low prevalence. Hyperparameters such as learning rate, batch size, and the number of epochs were kept consistent across models. However, architecture-specific hyperparameters were optimized independently for each model using Optuna.

The baseline Transformer followed the classic structure of a multi-head self-attention mechanism with positional encodings, while the RoFormer utilized rotary embeddings to encode positional information. The proposed variant replaced absolute position-based rotation with a mechanism that leverages relative time differences within each sequence, aiming to capture dynamic temporal dependencies more effectively.

### C. Results

Results demonstrate that the proposed variant ReDRE achieved the highest AUC-ROC of 0.740, outperforming both the RoFormer and the Transformer model, which attained AUC-ROC scores of 0.7288 and 0.7286, respectively. In all three models, the relative time gap between transactions is included, but in the standard RoFormer and Transformer, this gap is treated merely as another input feature. By contrast, the proposed approach ReDRE integrates these temporal intervals directly into the encoding process, yielding a more tailored representation of transaction timing and ultimately enhancing fraud detection performance.

The following table shows a summary of the results obtained for the models.

| AUC-ROC Scores | | |
|---|---|---|
| *Traditional Transformer* | *RoFormer* | *RoFormer ReDRE* |
| 0.7286 | 0.7288 | 0.740 |

In Fig. 1, we can see that while all three models exhibit similar behavior, RoFormer ReDRE demonstrates slightly superior performance across most of the AUC-ROC curve. Although the overall behavior of all models remains comparable, RoFormer ReDRE achieves a marginally higher classification ability, suggesting improved handling of temporal dependencies in transactional fraud detection.

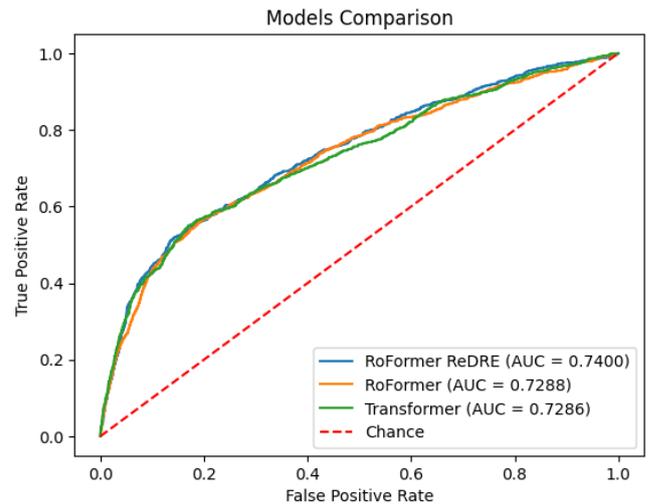

Fig. 1. Curve AUC-ROC for the Transformer, RoFormer and RoFormer ReDRE models.

## IV. CONCLUSIONS

In this work, we evaluated the performance of three Transformer-based architectures for fraud detection: a traditional Transformer, a RoFormer utilizing rotary position embeddings, and a proposed modified RoFormer variant that use rotary embedding based on relative distance between the events ReDRE. In both the traditional Transformer and the standard RoFormer, the relative time gaps between transactions are incorporated as separate input features within the feature vectors. In contrast, the proposed RoFormer variant integrates these relative time intervals directly into the input encoding process.

The experimental results demonstrated that the proposed model achieved the highest AUC-ROC of 0.740, outperforming both the RoFormer (0.7288) and the traditional Transformer (0.7286). While the overall behavior of all models remains similar, RoFormer ReDRE demonstrates a slightly superior classification capability, indicating enhanced management of temporal dependencies in transactional fraud detection.

To improve the obtained results and achieve a better prediction value, it is essential to enhance data preprocessing procedures. This includes refining feature engineering, handling missing values more effectively, optimizing data normalization techniques, and ensuring high-quality input data to improve model performance and robustness. These would be the next steps to effectively implement this technique for productive use in fraud prevention.

Finally, we can conclude that incorporating angle rotation using ReDRE enhances the characterization of time series data within a Transformer, broadening the applicability of this solution across a wide range of problems. Furthermore, this approach is generalizable to various types of event distances, allowing for greater flexibility in modeling sequential dependencies. By explicitly representing event distances, the model can more effectively capture and leverage the dynamic dependencies inherent in transactional data, ultimately improving fraud detection performance and robustness.